\def\year{2021}\relax
\documentclass[letterpaper]{article} %
\usepackage{aaai21}  %
\usepackage{times}  %
\usepackage{helvet} %
\usepackage{courier}  %
\usepackage[hyphens]{url}  %
\usepackage{graphicx} %
\usepackage{natbib}  %
\usepackage{caption} %
\begin{document}

\title{Model-agnostic fits for understanding information seeking patterns in humans}
\author {
        Soumya Chatterjee\textsuperscript{\rm 1}\thanks{Work done while at Google Research India},
        Pradeep Shenoy \textsuperscript{\rm 2}\thanks{Corresponding Author}\\
}
\affiliations {
    \textsuperscript{\rm 1} Indian Institute of Technology Bombay \\
    \textsuperscript{\rm 2} Google Research India \\
    soumya@cse.iitb.ac.in, shenoypradeep@google.com
}

\maketitle

\begin{abstract}
\begin{quote}

In decision making tasks under uncertainty, humans display characteristic biases in seeking, integrating, and acting upon information relevant to the task. Here, we build upon carefully designed experiments, and data collected at scale~\cite{Hunt2016}, that measured and catalogued these biases in aggregate form. We design deep learning models that replicate these biases in aggregate, while also capturing individual variation in behavior. A key finding of our work is that paucity of data collected from each individual subject can be overcome by sampling large numbers of subjects from the population, while still capturing individual differences. We predict human behavior with high accuracy without making any assumptions about task goals, reward structure, or individual biases, thus providing a model-agnostic fit to human behavior in the task. Such an approach can sidestep potential limitations in modeler-specified inductive biases, and has implications for computational modeling of human cognitive function in general, and of human-AI interfaces in particular.

\end{quote}
\end{abstract}

\section{Introduction}

A rich tradition of interdisciplinary research in cognitive science, economics, and computational modeling has studied humans' apparent suboptimality in decision-making tasks of various kinds (see, for instance, \citet{Gilovich2004}). In particular, humans appear to deviate significantly from task-optimal behavior in a fundamental aspect of decision-making under uncertainty: seeking and integrating relevant information in service of decision accuracy.  For instance, research suggests that instinctive ``approach'' and ``avoidance'' tendencies can interfere with learning action-reward  associations~\cite{Mkrtchian2017,Huys2011}, and prevent subjects from seeking balanced information~\cite{Hunt2016}; further, the integration of observed evidence and subsequent decision-making also appear to be influenced by irrelevant variables such as desired outcomes~\cite{Gesiarz2019} or framing of the question~\cite{Hunt2016,Kahneman1979a}. Modeling and understanding the computational provenance of such behavior is fundamental to the study of cognitive science; it may also underlie quintessentially human qualities such as optimism bias~\cite{Sharot2011}. The study of information seeking \& integration is equally relevant to a wide range of applications involving human-computer or human-AI interactions -- examples include preventing ``filter bubbles'' and polarization in recommender systems and media consumption~\cite{Pariser2011}, and efficacy of digital assistants or health \& wellness tracking software.

In addition to clever experimental design to empirically demonstrate biases~\cite{Hunt2016}, research also aims to understand the underlying computations driving these biases. Typically, under the constraints of a specific task, human behavior is hypothesized to arise from carefully specified models, whether \textit{computational}~\cite{Marr1982} such as reward maximizing using reinforcement learning, bayes-optimal under a specific uncertainty model, etc., or \textit{algorithmic}, i.e., in terms of mechanism, such as drift-diffusion processes~\cite{Ratcliff2008} or neuronal circuits performing predefined computations~\cite{Ma2006}.

A critical commonality in much of the literature is the strong {\em inductive bias} implicit or explicit in proposed models,  which are validated against each other in terms of quality of data fit, or  via correlational analyses against task-extrinsic  data such as personality traits or diagnosis. Two major challenges to this  approach are (a) the extreme paucity of data for subject-specific model fits to data, and (b) limitations of modeler-specified inductive biases in capturing the actual, underlying computations in humans.  An increasing trend towards large-scale experimentation in naturalistic settings~\cite{Brown2014} partially alleviates data paucity by enabling data collection from larger numbers of participants. These trends present an opportunity for developing new modeling \& analysis techniques that can effectively leverage the availability of large-population data.
 
\begin{figure*}[!h]
\centerline{
\includegraphics[height=1.9in]{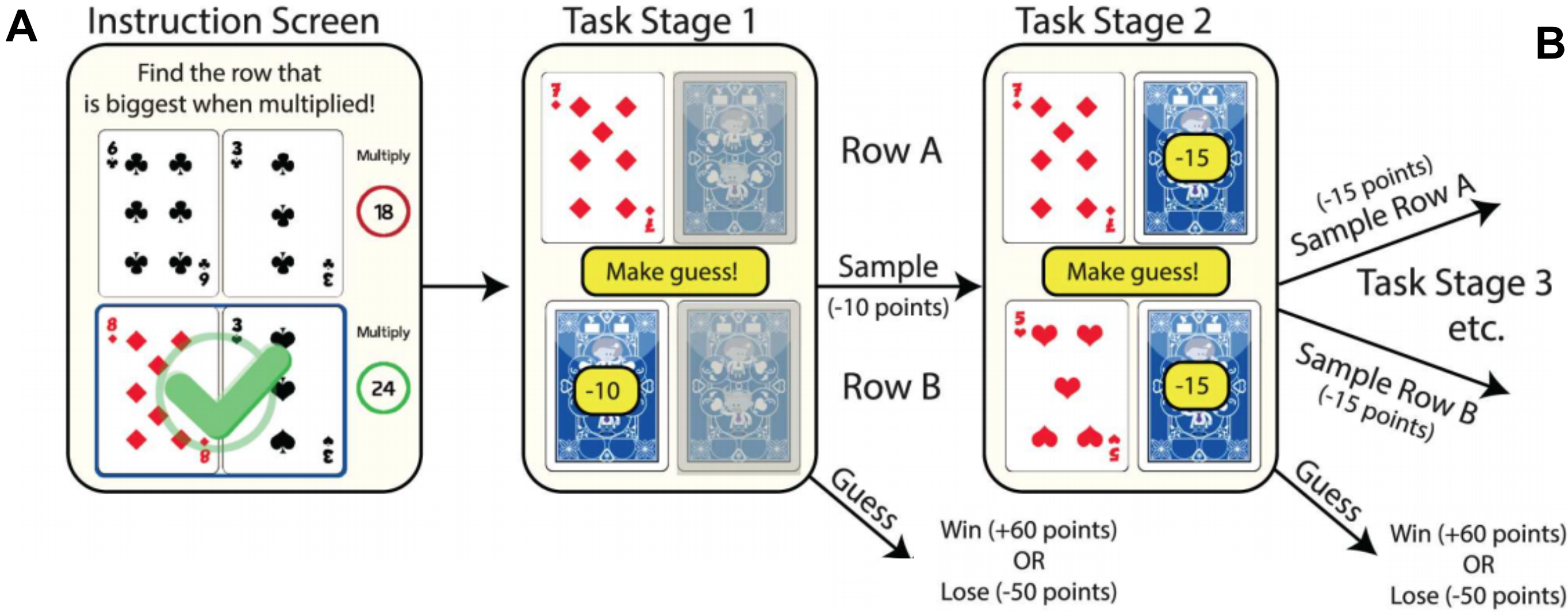}
\includegraphics[height=1.9in]{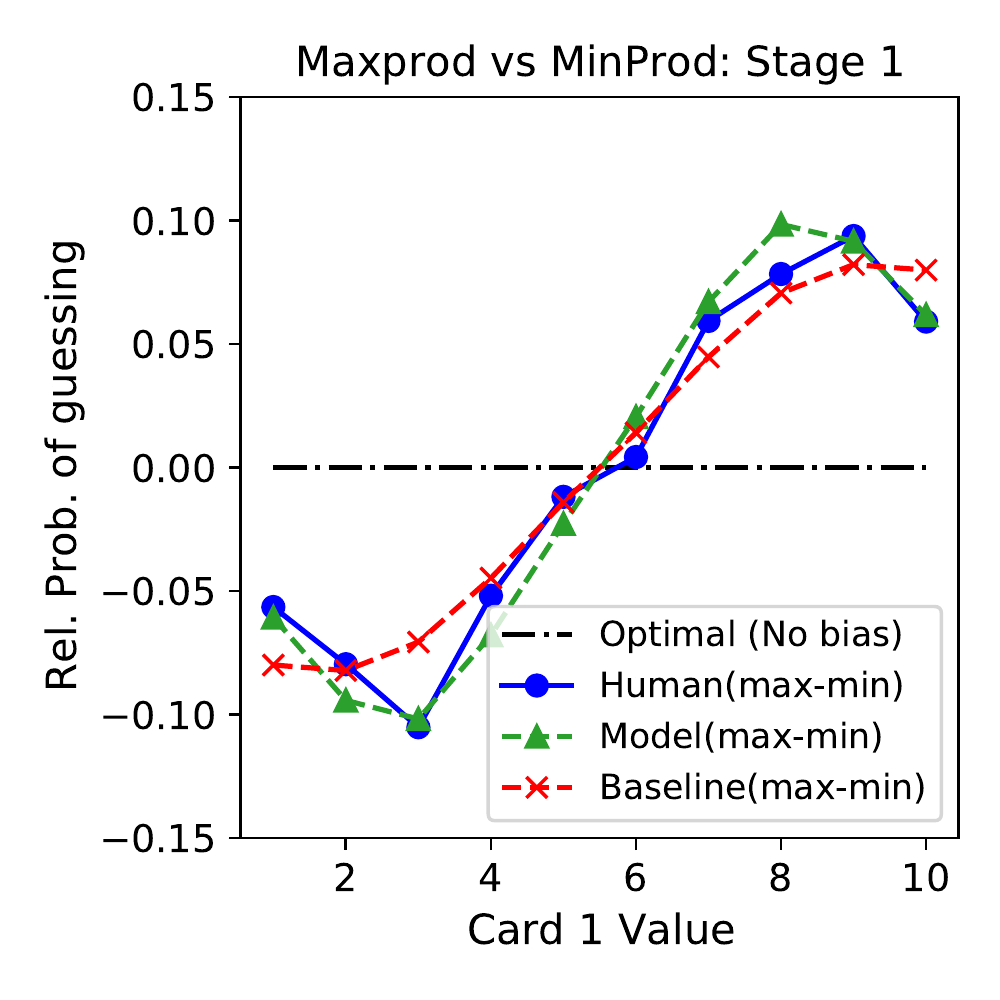}
}
\caption{Task structure \& biases: Panel A shows the multi-step decision-making task studied in humans. Subjects at each step choose between opening a card from an offered row, or deciding which row has the maximum product of cards (MaxProd, similarly with minimum product, MinProd). The cost of sampling, and rewards/penalties for correctness, create a trade off between reward and accuracy. Panel B shows the framing effect in human behavior and our model. The Y axis shows the difference between Maxprod and MinProd, in the probability of guessing after seeing 1 card. This difference is nonzero (i.e., framing influences guess probability), and modulated by card value (guessing more likely in MaxProd with large first card). Both the baseline \& DNN models capture this broad trend. See text for details. (Panel A adapted from~\citet{Hunt2016})} \label{fig:task-biases}
\end{figure*}

In this paper, we propose a deep learning approach for fitting human behavior, in very sparse data scenarios, by leveraging large subject populations and simultaneously learning population- and individual- level model fits in a single learning process. We apply our methods to a recent large study cataloguing biases in human information seeking, integration, and decision-making~\cite{Hunt2016}, and demonstrate the following results:
 
\begin{itemize}
    \item {\bf model-agnostic} fits to human behavior: By using a multi-task DNN architecture that reflects the \textit{structure} of the behavioral task, but not the goals or rewards, we recover subjects' action policies purely by replicating  behavior. 
    \item {\bf computational fits to human biases:} Our models capture empirical biases in the task at population level, and, to a substantial degree, at individual level.
    \item {\bf subject-specific} fits in sparse data scenarios: We capture behavioral variation across subjects in the task, using as few as 6 trials per subject, by pooling together data from a large number of subjects.  We show through experiments that increasing the subject pool directly improves the ability to fit policies to any one subject's data, without any additional data from that subject.
\end{itemize}

We believe  our approach is broadly applicable to a range of experimental paradigms in cognitive- and neuro-sciences, opening up the possibility of new scientific discovery through large-scale data modeling techniques. Although our models are purely predictive in nature, i.e., they are not intended as an explanation of underlying cognitive processes in the task, we hope to establish a compelling {\em upper bound} on predictive power in the task, to serve as a benchmark for other, more cognitively inspired, inductive biases. Furthermore, decision policies implicit in our networks can be explicated using a host of recent techniques for interpreting learned models--see for example~\cite{NDPS18, symbolic20}--this is an active area of  future work. Finally, for a range of applications in human-AI interactions, predictive power and ability to capture biases are valuable properties independent of interpretability.

The rest of the paper is organized as follows: We first describe the behavioral task from~\cite{Hunt2016}, including their empirical findings of biased decision-making. We describe our modeling approach, and show that it recovers population-level biases from the empirical findings. We show that our model captures individual variation in the task, and achieves this result via learning compact, subject-specific embeddings. We show that increasing the number of subjects indeed improves ability to model individual subjects' behavioral policies. We conclude with a brief summarization of the related work in cognitive science \& machine learning, and a discussion on future work.

\section{Task \& empirical results}
\subsection{Information-seeking and decision making}

We model data from a cognitive experiment conducted on a large-scale mobile-phone-based experimental platform~\cite{Brown2014} to probe decision-making under uncertainty~\cite{Hunt2016}\footnote{All data were released by authors as part of the publication~\cite{Huntdata} under the CC0-1.0 license.}. Briefly, subjects played a multi-stage guessing game starting with 4 cards of value 1-10, face-down in a 2x2 configuration. In some games, the objective was to guess the row with the maximum product (MaxProd); in other games, to guess the minimum product (MinProd). Note that given only two rows, these tasks are opposites of each other; this allowed for probing the effect of \textit{task framing} on decision-making. At each stage, the subject could choose whether to sample from a pre-specified row, or to guess the answer and terminate the game. By controlling which row the subject could sample from, the experimenters could check for biases in information seeking based on current information and task framing. Each stage had an increasing cost of sampling (0,10,15,20 points), and the final answer was awarded +50/-60 points based on correctness. The task workflow is presented in Figure~\ref{fig:task-biases}A.

The dataset contains data from 13915 subjects who've played the ``MaxProd'' game and 13242 subjects who've played the ``MinProd'' game, containing an overlap of 3230 subjects who've played both games\footnote{The asymmetry is due to the design of the original experiment, in which each participant played 2 games from a large set (e.g., MaxSum, MinSum, etc.; here, we focus for simplicity on only the MaxProd and MinProd games, and can therefore use only part of the original data.}. Typically, subjects complete at least 1 block (11 trials) of each condition, although a small number of subjects have completed more blocks of trials with an average of 14.5 trials per subject.

\begin{figure}[!t]
\centerline{
\includegraphics[width=0.47\textwidth]{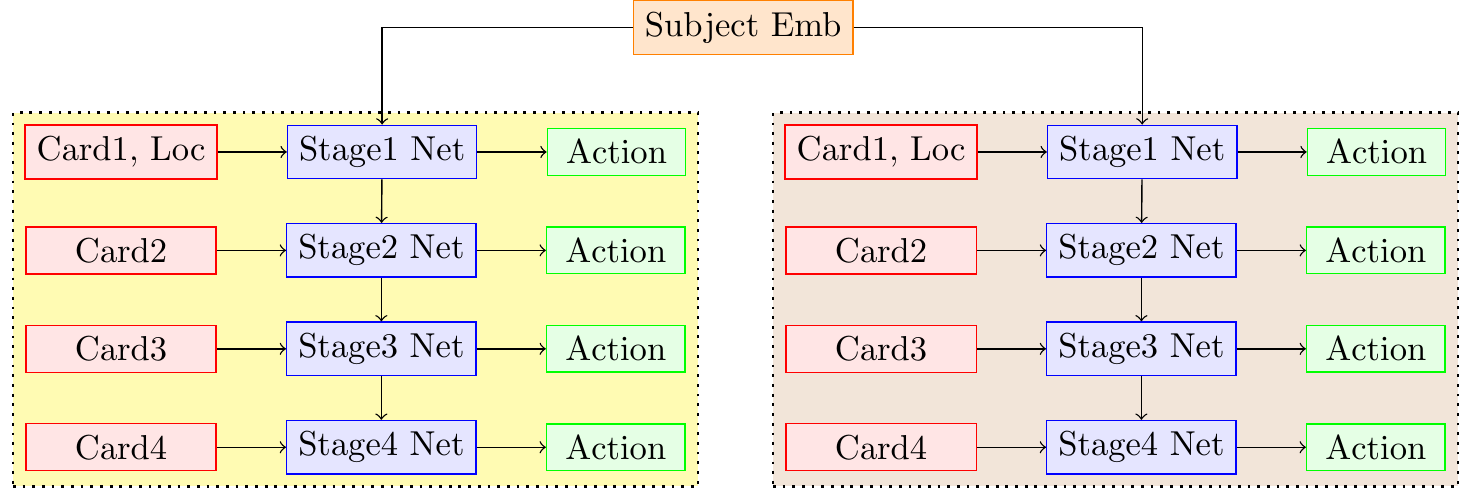}
}
\caption{DNN Model Architecture. We propose a multi-stage multi-task network with learned embeddings to model human behavior. The two parallel networks represent alternate framings of the task (MinProd vs MaxProd), while each stage represents a within-trial stage of decision making (sample vs guess). Both networks share a single, learned embedding for each subject in the data. See text for more details.}\label{fig:model}
\end{figure}

\subsection {Empirical findings of approach biases}

\begin{figure*}[!t]
\centerline{
\includegraphics[width=6in]{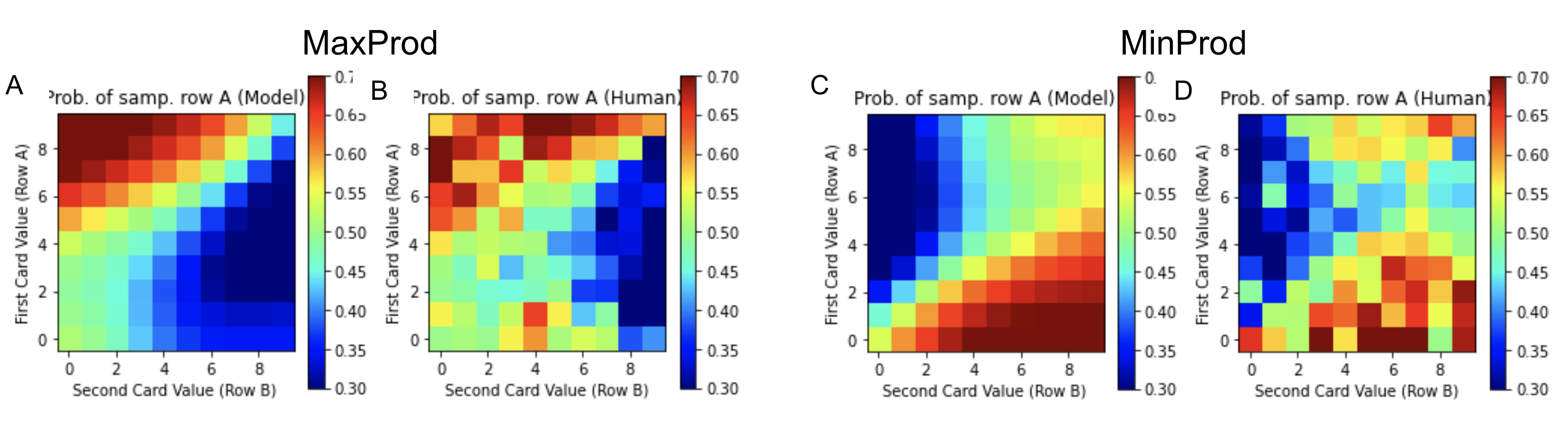}
}
\caption{Approaching the positive: After having seen one card in each row, the probability of sampling a particular row should be independent of whether the task is MaxProd, or MinProd. However, subjects (Figures B, D) show markedly different sampling patterns, preferring to sample from the row likely to be the answer. The Multi-DNN model reproduces these biases (A, C). See text for details.} 
\label{fig:biases2}
\end{figure*}

Some key findings in the experiment were as follows:

\begin{enumerate}
   \item {\bf Framing bias:} The framing of the problem influenced subject behavior (Figure~\ref{fig:task-biases}B). Subjects showed differential likelihood of guessing, modulated by card value, based on whether the goal was to find maximum product, or to find minimum product, even though these goals are conceptually identical in a 2-choice game,
   \item \textbf{Approaching the positive:}  Subjects were more likely to accept a sampling option if it was from the row they were eventually going to select; this suggests a tendency to look for confirmation of choice, rather than to maximize information content, in sampling decisions (Figure~\ref{fig:biases2}). 
   \item  \textbf{Rejecting the unsampled:} Subjects were less likely to choose a row as answer if they had passed on the option to sample from it (Figure~\ref{fig:biases3}); this strengthens the observation from the previous bias that sampling decisions are correlated with the current hypothesis / final choice. 
\end{enumerate}

\citet{Hunt2016} fit a softmax choice model to subjects' behavior at each stage, at population-level (i.e., pooling data from all subjects), with the following inputs a) mean-subtracted card value of open cards, b) a free parameter for each kind of observed bias. They show that the population-level model captures behavior in the task, as well as the observed approach-related biases, at population level.

\section{Model-agnostic DNN fits to behavior}

\begin{figure}[!t]
\centerline{
\includegraphics[width=2.5in]{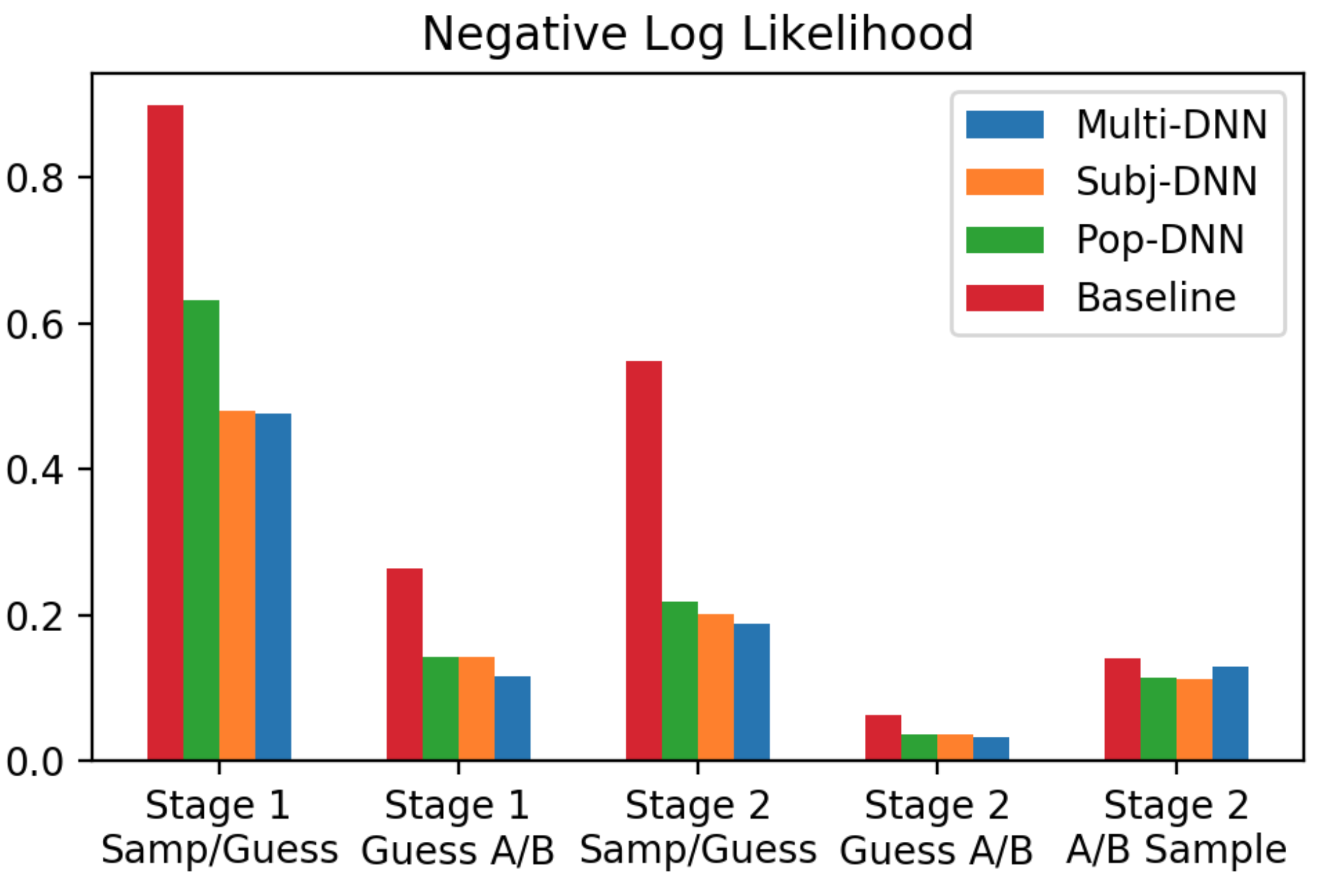}
}
\caption{Log Likelihood of proposed models. The figure shows that our proposed models (*-DNN) better model the likelihood of data overall, compared to baseline; further, the subject parametrization and multi-task architecture cumulatively improve model fit. See text and Figure~\ref{fig:task-biases} for details. }\label{fig:loglik}
\end{figure}

We propose a multi-task DNN architecture with shared subject-specific parameters (see Figure~\ref{fig:model}), with a separate network for each of the two symmetric tasks ``MaxProd'' and ``MinProd''. Each task network has separate layers per task stage, with inputs corresponding to the newly seen card value and its position, and outputs corresponding to the three choices: sample/guess, row to sample (if sampling), and row to guess (if guessing). A shared \textit{embedding layer}  represents each subject by a learned embedding, and can help capture individual differences in the task. We use the multi-task approach to integrate behavior from alternate framings of the task (MaxProd/MinProd), and capture behavioral asymmetry at an individual level using the subject parameters. 

Our goal is to predict human behavior in a \textit{model agnostic} manner, i.e., without assuming specific functional forms or policy family. Indeed, our results show that the network can accurately capture human behavior, and its biases, without explicit encoding of task goals anywhere in the network or training. We believe that providing a \textit{lightweight parametrization} for subjects can capture individual variability, while sharing the \textit{policy family} (i.e., the multi-task DNN) allows leveraging of pooled data across subjects. Results show that this simple parametrization does capture variation in individual behavior to a great degree, while still reflecting the population-level biases observed in the data.

\subsection{Model architectures \& training}

Each stage of a task network takes as input the following: sampled card value (normalized numeric), offered sample row for the next card (binary), latent state from previous stage (dimension 10, for stages 2-4), and subject embedding (dimension 2, for stage 1). The stage network transforms these inputs in the following manner: a) 2 fully connected layers of dimension 10 to produce the next hidden state, b) 2 additional single fully connected layers on top of this hidden state, one each for producing the decision outputs of that state.  The decision variables are, respectively, whether to guess or sample, and which row to select as final answer. Hidden state from previous layers are also passed to the output layers, alongside hidden state from current layer. All activation functions are $tanh()$. Inputs and outputs irrelevant to the stage are dropped from the architecture. Training is only performed on  portions of the network corresponding to activated decision pathways in behavior.

Data were split into 60/40 {\em per subject} for training and validation--6-7 trials of training data per subject per task (2x for Multi-DNN). Training used Adam optimizer with learning rate 0.003 and batch size 256, for 30 epochs with early stopping. We varied the width and number of hidden layers in our experiments, but found no significant variation in the results.
We experimented with the following models:\\
\noindent{\bf Baseline:} Population-level model proposed by \citet{Hunt2016}, who fit a separate softmax model, for each decision stage, for each task, at population grain.\\
{\bf Population DNN:} We train a single DNN model (one column of Figure~\ref{fig:model}) for each of the tasks MaxProd and MinProd, using data pooled across all subjects.\\
{\bf Subject-specific DNN:} Similar to Population DNN; however, we also learn subject-specific 2-d embeddings. \\
{\bf Multitask DNN:} We train the full model shown in Figure~\ref{fig:model}, which includes behavioral data from both tasks, and a shared subject embedding learned from that data.

We call the DNN models Pop-DNN, Subj-DNN and Multi-DNN respectively in the following sections.

\subsection{Decision-making biases at population level}

Figure~\ref{fig:loglik} shows the average negative log-likelihood of all choice data in the validation trials, under each of the proposed models, for the MaxProd task. The data show that each of the DNN models provide significantly better fit than the Baseline model; this is precisely because our model does not presuppose a specific computational objective, but instead attempts to faithfully reproduce observed behavior. Further, the Subj-DNN shows substantial additional improvement, suggesting that modeling individual variation is important in providing an accurate account of the data. Finally, the Multi-DNN model shows modest further improvements -- we note here that although the Multi-DNN benefits from additional variety of data (behavior from two different tasks), it is trained on far fewer subjects, 3230 opposed to 13915, since fewer subjects have performed both tasks. 

In addition to log likelihood, our model faithfully captures population-level \textit{biases in information seeking} in the data, as observed by \citet{Hunt2016}. In Figure~\ref{fig:task-biases}B, we see that the model captures the {\em framing effect} bias: there is a difference, between MinProd and MaxProd tasks, in the likelihood of making a final choice after seeing only one card. This difference is suboptimal in the particular setting since finding the maximum product among two rows is equivalent to finding the minimum product. The figure shows that both the Baseline model and our Multi-DNN model capture this bias in human behavior. (Pop-DNN and Subj-DNN also capture this aggregate trend; data not shown).

Figure~\ref{fig:biases2} shows the {\em approaching the positive} bias. Panels B, D contrast the behavior of human subjects on MaxProd \& MinProd tasks respectively, after one card from each row has been shown. The intensity shows the likelihood of sampling from the first row, as a function of the two card values seen so far (\textit{y} \& \textit{x} axes, respectively). From an optimal decision-making perspective, the two figures should be identical; however, from the patterns it is apparent that subjects are more likely to select from the row that they already expect will be the final answer. Panels A, C show that our Multi-DNN model faithfully captures this asymmetry.

\begin{figure}
\includegraphics[width=\linewidth]{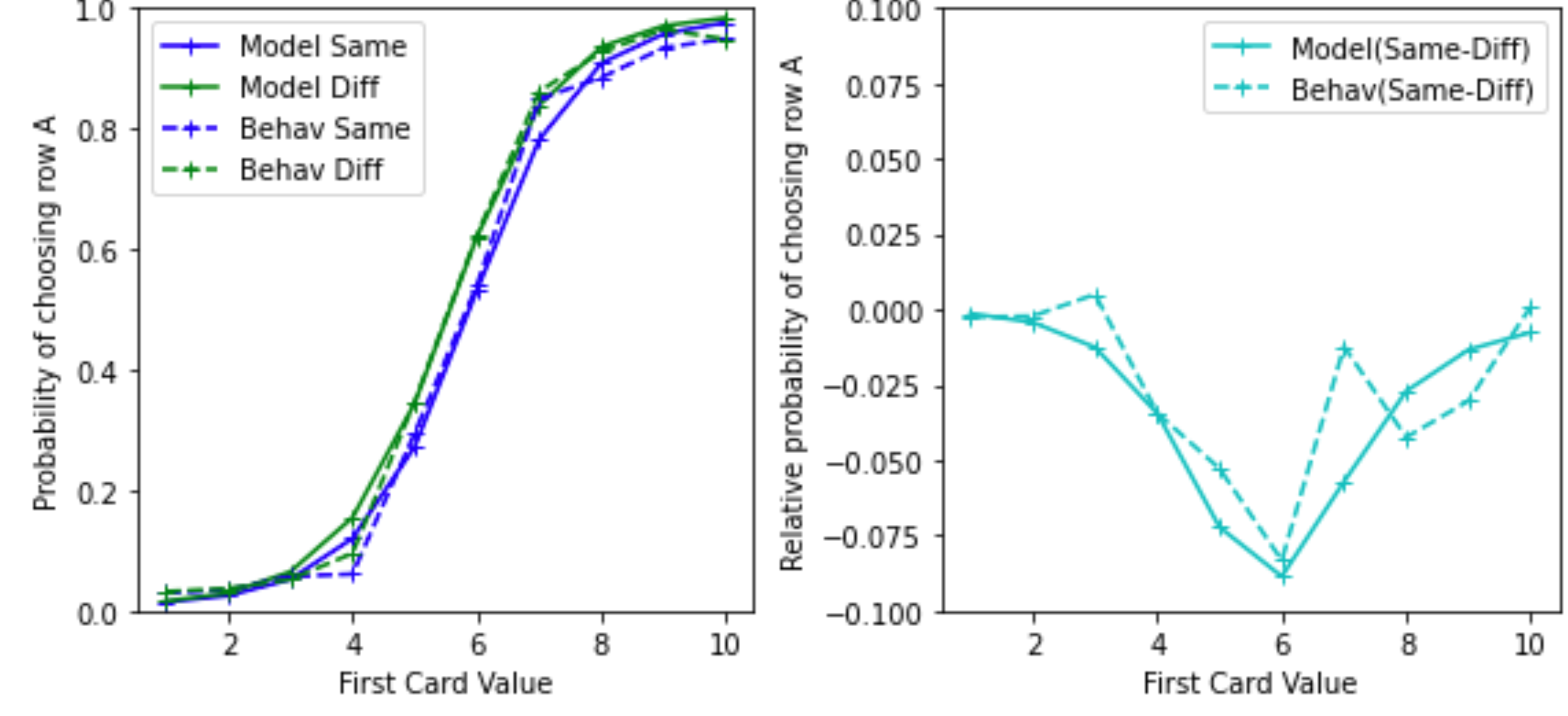}
\caption{Rejecting unsampled options: Panel A shows probability of guessing ``row A'' as answer, on trials where subjects guessed immediately after the first card. Trials were split by whether the next offered sample was from the same row, versus from the different row.  Although these two curves should be identical in an optimal model (i.e., which row was offered for sampling should not correlate with one’s decision), we see a clear difference; more readily apparent in the difference between the curves (panel B). The model captures this bias. See text for details.} \label{fig:biases3}
\end{figure}

In Figure~\ref{fig:biases3}, we show that the model captures the {\em rejecting the unsampled} bias: in the MaxProd game, after choosing to not sample, the subjects showed different probabilities of selecting row 1 as final answer, when comparing trials where samples were offered from row 1 vs from row 2. This demonstrates a correlation between whether to sample from a particular row, and whether that row is the current hypothesis; in particular, subjects appear to sample more from current hypothesis (confirmatory behavior) than from the other row, even when the latter could be more informative. 

\subsection{Capturing individual variation in the task}

We trained the Multi-DNN model, along with subject embeddings, on pooled training data from all subjects, and simulated the model's choices on the validation data for each subject. We calculated performance metrics at a subject grain (\# moves, accuracy, and total score) from  model simulations, as well as from human behavior.

\begin{figure*}[!t]
\centerline{
\includegraphics[width=6.5in]{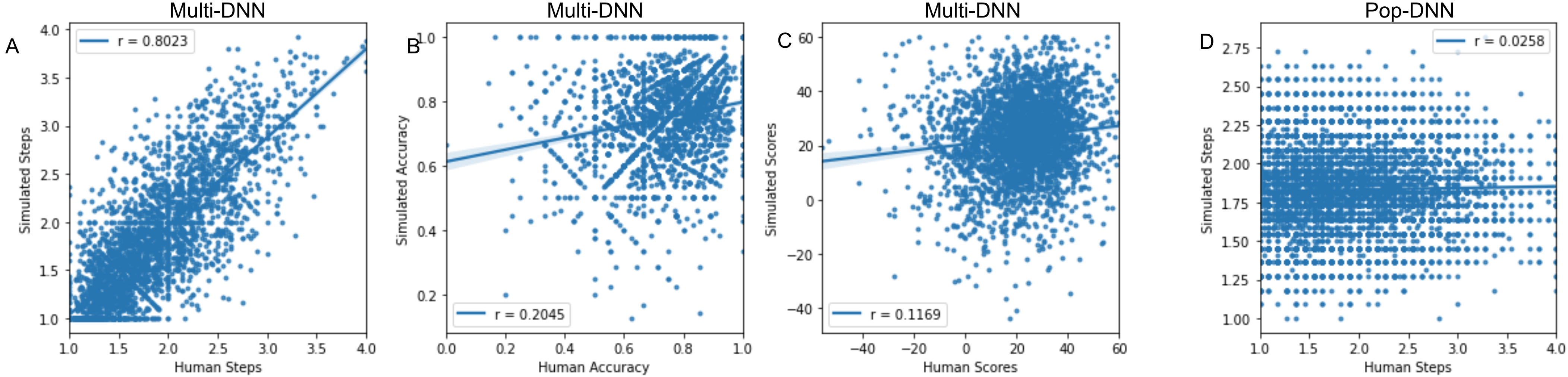}
}
\caption{Individual variation in behavior. We simulated the learned model, with subject embeddings, on each subject's held-out data, and compared task measures on simulated data against actual behavior. Results show that variations in average \# moves (panel A), and accuracy (panel B), are captured fairly faithfully using as little as 13 trials per subject. Total score (panel C) also shows significant correlation, although of lesser magnitude (pearson correlations, Panels A-C:  p-values $<$ 1e-10, Panel D: n.s., all figures sample size=3230 subjects).} \label{fig:subjmetrics}
\centerline{
\includegraphics[width=\linewidth]{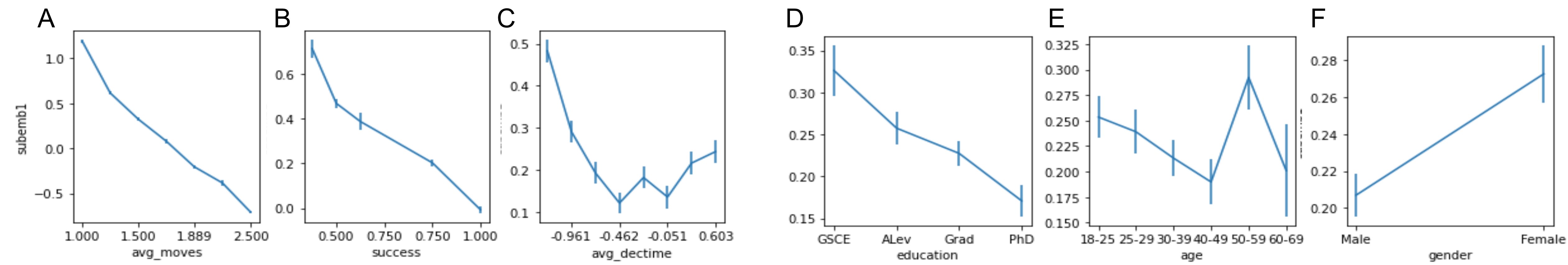}
}
\caption{Subject Embeddings capture behavior \& demographics. Panels A,B,C bucket subjects by decile of average \# moves, success rates, and (log) decision times respectively, and plot the mean$\pm$SEM of the subject embeddings for subjects in each bucket. We see that subject embedding values are significantly different across buckets, and are the source of the model's ability to capture behavioral variation. Panels D-F show distinguishable differences in embedding values across subject education levels, age buckets, and gender. (Data shown for embedding dimension 1 only, for space; dimension 2 qualitatively similar.)} \label{fig:emb_corr}
\end{figure*}

Figure~\ref{fig:subjmetrics}A-C shows that our model simulations capture significant variation in the observed human behavior (sample size = 3230 subjects, all correlations p $<$ 1e-10, i.e., highly statistically significant). Average steps are highly correlated (r=0.80), while accuracy and scores show lower correlation values.\footnote{We note briefly that the latter two measures suffer from severe estimation issues --  4-8 validation trials per subject are not sufficient to accurately calculate a subject's average behavior, due to the substantial variation introduced by the specific card layouts in the trials. As a consequence, the correlation between model and human are also systematically under-estimated} This result demonstrates that the subject embeddings in Multi-DNN effectively parametrize individual behavior, using very small amounts of data.

In order to  further rule out the possibility of this result being simply due to noise, e.g., stochasticity in the specific trials included in each subject's validation data, we simulated Pop-DNN on the same validation data. Figure~\ref{fig:subjmetrics}D shows that Pop-DNN predictions are uncorrelated (r=0.025, n.s.) with human behavior (data shown only for average \# moves); this further underscores the role of the subject embedding in capturing the range of individual variation.

\subsection{Subject embeddings as parametrizations of decision policy}

As seen in the model architecture (Figure~\ref{fig:model}), the key to our model's ability to personalize the decision policy to each individual subject, is the \textit{subject embedding}; in our case, a learned vector of dimension 2. As secondary evidence that the embeddings capture variance in behavior, we correlated the value of the learned parameters against the same behavioral measures (Figure~\ref{fig:emb_corr}A-C).\footnote{This and subsequent analyses focus on dimension 1 of the 2-d embeddings, for simplicity and space; results are qualitatively similar for dimension 2. We found that 2-d embeddings performed marginally better than 1; additional dimensions did not further improve performance.} Figure~\ref{fig:emb_corr} shows that, indeed, the learned embeddings do contain information about subjects' performance, since the average embedding values are statistically different across buckets of \#moves (A), and similarly, choice accuracy (B), as well as decision time (C). The latter is an interesting result, suggesting that decision time, typically related to subjective uncertainty about choice, is also captured by the Multi-DNN model, despite having no access to response time data during training. Further, since the collected data included demographic information (age, gender, education) which were extrinsic to the experimental setup, we tested whether the learned embeddings contained information regarding demographic data. In Figure~\ref{fig:emb_corr}D-F, we see that the subject embeddings show statistically significant differences in value across age groups, and across educational levels. As noted in \citet{Hunt2016}, these demographic dimensions are already associated with behavioral differences in the task; hence it is not surprising that the embeddings (which are predictive of individual task performance, see Figure~\ref{fig:subjmetrics}) also show differences across these dimensions. Still, it is additional validation that the learned subject embeddings capture some aspects of individual traits both within the task and beyond.

\subsection{Empirical tradeoffs in sample complexity}

We examined the relationship between quality of fit for each individual subject, and the {\em number of subjects} in our training pool, to check whether larger subject pools indeed help learn better predictive models. In this experiment, we measured simulation accuracy on held-out trials for a fixed cohort of subjects designated as ``test-subjects''. For training, we used data from these test-subjects, plus an increasing number of additional subjects. For validation, we only calculated metrics on the test-subjects. In this manner, we can examine the influence of adding {\em extrinsic data} (i.e., data from other subjects) to the quality of fit, and thereby the capturing of individual variation, of a fixed set of test-subjects.  

\begin{figure}[!t]
\centerline{
\includegraphics[width=\linewidth]{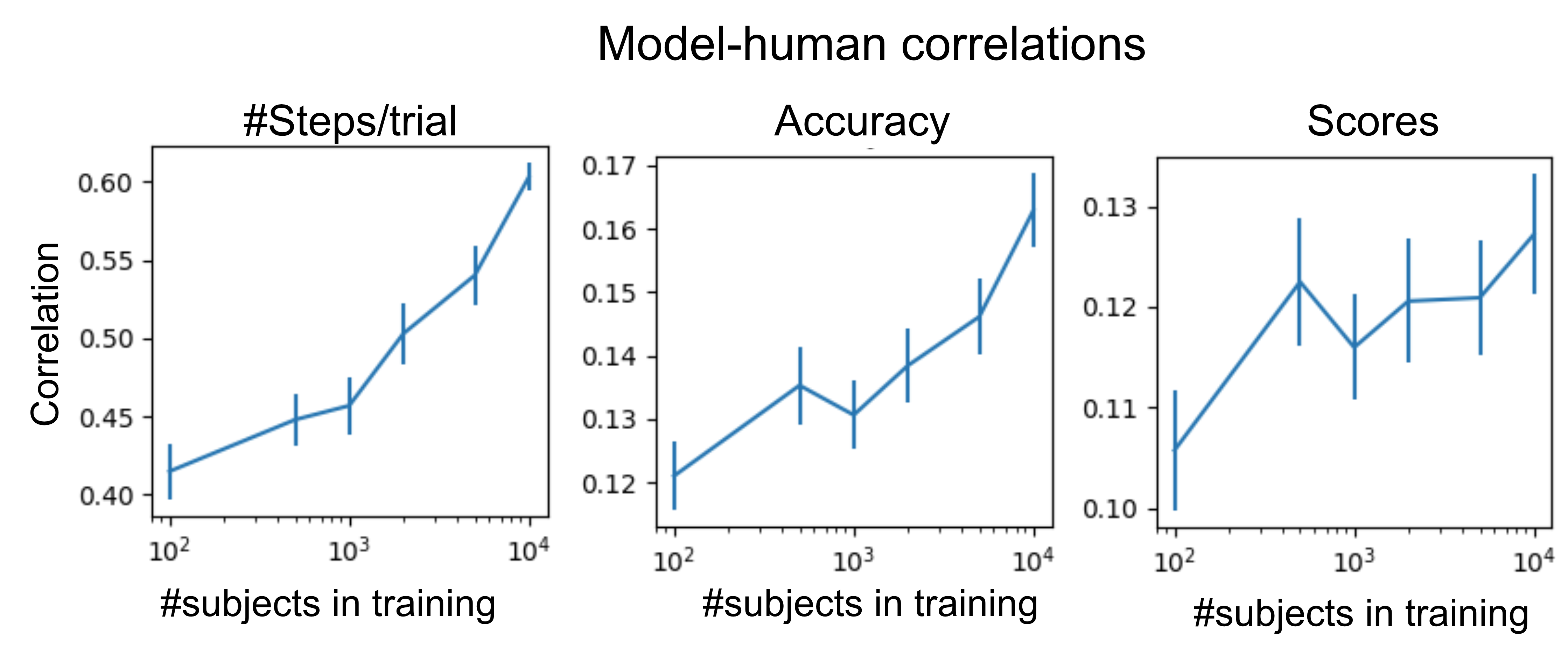}
}
\caption{Empirical sample complexity. The figure shows the correlation between model-predicted behavioral measures, and actual behavior, for the Subj-DNN model on the MaxProd task. The correlations are calculated over a fixed set of 100 subjects included in both training \& validation data, as we add additional subjects to the training data alone. Curves represent mean$\pm$SEM of correlation over 50 runs of the evaluation. Panels A-C show correlations for average \# moves, accuracy, and average scores, respectively.} \label{fig:sampcomp}
\end{figure}

For each instance of the training subject pool, we trained the Subj-DNN model on subjects' behavior in the MaxProd task\footnote{We used MaxProd task / Subj-DNN alone, since the data contained a greater number of subjects who had performed this task. The Multi-DNN uses subjects who have performed both MaxProd and MinProd tasks, a significantly smaller pool.}. Each subject in this experiment had 6-7 trials of training data and 4 trials of validation data. We simulated the trained model on the validation trials {\em for the 100 test-subjects alone}, and calculated per-subject behavioral metrics as above. We calculated the correlation at subject-grain between model simulations and actual human behavior from the validation data. This calculation was performed for each training pool, and repeated 50 times to identify average trends if any.  Figure~\ref{fig:sampcomp} shows the results of this experiment; each panel shows the correlation (mean $\pm$ SEM over 50 runs) between human and model behavior at subject grain over the test-subjects.  As the number of subjects in the training data pool increases, the correlations steadily increased. Note that the only manipulation is {\em the addition of subjects into the training pool}; for the test-subjects, the available data for training and validation is unchanged. 

This result provides strong support for our main hypothesis: we compensate for the lack of {\em per subject data} through the use of (a) a parametrized, shared model across subjects, and (b) a large pool of subjects in the training data. By adopting this approach, we show not only improved fits to individual behavior, but also the ability to {\em better capture individual variation} in the task (the correlations measured in Figure~\ref{fig:sampcomp}).

\subsection{Multi-DNN captures biases at subject grain}

Finally, although Figure~\ref{fig:subjmetrics} demonstrates that gross measures of human performance at the task can be recovered at per-subject grain, the question remains whether the subtler behavioral idiosyncrasies in information-seeking (Figures~\ref{fig:task-biases},~\ref{fig:biases2},~\ref{fig:biases3}) are also captured at an individual level by our model, or if the model lacks sufficient sensitivity to capture such biases. Clearly, as seen in Figure~\ref{fig:biases3}, the behavioral differences are so slight that mere 4-8 trials of validation data we have per subject are not enough to even get an empirical measure of individual bias, for validating our model.

\begin{figure}[!t]
\centerline{
\includegraphics[width=2.7in,height=1.4in]{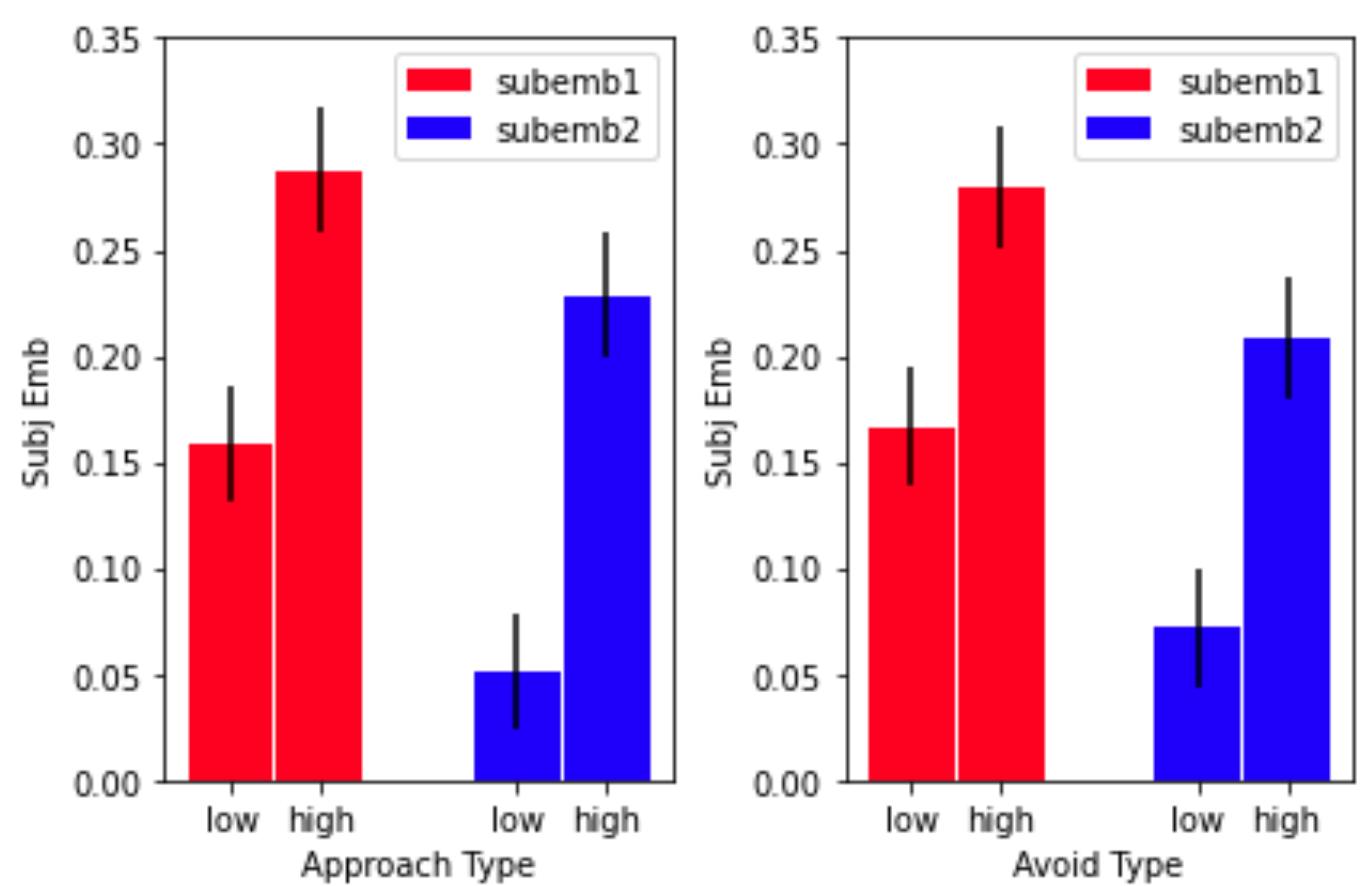}
}
\caption{Capturing biases at individual grain: The figure shows subject embeddings split into buckets by median value of ``approach'' and ``avoidance'' subjective parameters estimated on a separate task (see text for details). The bars represent mean$\pm$SEM of subject embedding values within each bucket. As seen in Panels A,B, both embedding values show clear difference across the low- and high-value buckets of the approach and avoid parameters respectively.} \label{fig:approachavoid}
\end{figure}

Instead, we follow \citet{Hunt2016} and use data extrinsic to the task in order to establish validity. The data from \citet{Huntdata} includes, for each subject, a parameter estimate of ``approach'' and ``avoidance'' tendencies calculated on a different, gambling-based task~\cite{rutledge2015dopaminergic}. Figure~\ref{fig:approachavoid} shows the average subject embedding value for each of these extrinsic parameters, split into ``low'' and ``high'' buckets via the median value. The bars represent mean$\pm$SEM for each bucket. As can be seen, the subject embeddings show clear differences in the low and high buckets for both the approach and avoid parameters, suggesting that embeddings capture a more general subject-specific aspect of approach behavior than reflected by the specific task.

\section{Related Work}

\subsection{Uncertainty \& information seeking}

Since our focus here is on developing general-purpose modeling tools \& methodology, we do not provide a comprehensive review of the cognitive science literature on uncertainty in decision-making, and biases in information sampling tasks; instead, we provide a brief sketch of the related work to round out the understanding of the experiments discussed here.

A rich literature deals with humans handling uncertainty in the form of risk (stochasticity of outcome~\cite{Ellsberg1961}) and ambiguity (uncertainty with respect to world models / estimates of probabilities, or ``second order uncertainty''~\cite{Bach2011,Yu2005}), showing that both play a role in human decision-making. In particular, humans appear to be ambiguity-averse~\cite{Bach2011}, even if the ambiguity does not affect expected value of outcomes. In terms of information-seeking, well-documented behavioral tics such as confirmation bias~\cite{Nickerson1998} and optimism~\cite{Sharot2011}, suggest that in certain situations, humans are prone to \textit{selectively choosing} information on which to make decisions, rather than integrating all available information in a ``fair'' or unbiased manner. In other related work, there is substantial evidence of ``approach''/``avoidance'' behaviors in humans, where certain scenarios engender an automatic approach or avoidance action, even when the reward structure punishes such automatic behavior~\cite{Huys2011,Mkrtchian2017}. From a modeling perspective, much of the literature starts with these idiosyncrasies as a fact of human behavior (i.e., an inductive bias), and explores variants of reinforcement learning models incorporating free parameters for some of these tendencies~\cite{Hunt2016,Huys2011,Mkrtchian2017}. In our work, we show better model fits to data by using black-box architectures (i.e., by side-stepping any inductive bias); further, the behavioral policies (and individual variation) are distilled into a small number of subject specific parameters that could be used in downstream applications, e.g., for demographic analysis or diagnosis~\cite{Dezfouli2019b}. The question of \textit{explaining the subject's policies} is still open, however, and an interesting modeling challenge for interpretable machine learning, one we hope to explore in future work.

\subsection{Modeling}

Recent literature in cognitive science~\cite{Dezfouli2019a,Dezfouli2019b} has examined the use of DNNs for model-agnostic fits to human behavior. \citet{Dezfouli2019a} model a simple probability tracking task using LSTMs, instead of the classical reinforcement learning models popular in the literature. They show that LSTMs, being model agnostic, can capture arbitrary policies like deterministic switching irrespective of historical value,  policies that are actually seen in behavior but not captureable by simple value-based models such as Q learning. In subsequent work~\cite{Dezfouli2019b}, they learned disentangled embeddings for each subject using behavioral data. We build substantially on these promising initial drections with the following additional contributions: (1) application to a multi-stage sequential decision-making task, and recovery of idiosyncratic biases (``suboptimalities'') from behavior, and critically, (2) demonstration that pooling together large numbers of subjects can enable learning individual behavior from very small sample sizes (5-10 trials per subject).

From the machine learning perspective, several approaches overlap in spirit with our goal here: work in meta-learning~\cite{Finn2017, Garnelo2018} is interested in separating out the learning of a \textit{class of functions} from the learning of \textit{instances of that class}, and may broadly be applicable to tasks such as the one studied here. Our work is a key proof-of-concept that meta-learning is a useful tool in cognitive science; in future work, we will examine whether specific ideas from the meta-learning literature can be used to push our results further ahead.

\section{Discussion}

We presented a model-agnostic, multi-task approach for modeling human behavior in an information-seeking task. Our model recovered broad behavioral trends, including well-documented biases in the task~\cite{Hunt2016}, without access to task objective or reward structure during training. Crucially, we capture individual variation in the task using as little as 6-8 trials per subject for training, by leveraging data from a large pool of subjects. These results have significant implications for cognitive modeling efforts in general: (a) meta-learning approaches, such as our proposal, can sidestep the typical challenge of extreme data paucity in learning individual subject traits, (b) model-agnostic approaches can avoid limitations of inductive bias introduced by modelers, and capture data and action policies more faithfully.  Some natural next steps in this line of research: first, using other, more powerful meta-learning approaches for strengthening the predictive power of models, and second, developing {\em explainable policy approximations} to learned DNNs as a collaboration between machine learning and cognitive science experts. We hope to use such policy interpretations to better situate our findings in the context of theories of approach \& avoidance that form the bedrock of seemingly biased behavior in these tasks, and may help shed more light on the computational provenance of ubiquitous human tendencies such as confirmation bias \& optimism.

\bibliography{refs}

\begin{thebibliography}{24}
\providecommand{\natexlab}[1]{#1}
\providecommand{\url}[1]{\texttt{#1}}
\providecommand{\urlprefix}{URL }
\expandafter\ifx\csname urlstyle\endcsname\relax
  \providecommand{\doi}[1]{doi:\discretionary{}{}{}#1}\else
  \providecommand{\doi}{doi:\discretionary{}{}{}\begingroup
  \urlstyle{rm}\Url}\fi

\bibitem[{Bach et~al.(2011)Bach, Hulme, Penny, and Dolan}]{Bach2011}
Bach, D.~R.; Hulme, O.; Penny, W.~D.; and Dolan, R.~J. 2011.
\newblock {The known unknowns: Neural representation of second-order
  uncertainty, and ambiguity}.
\newblock \emph{Journal of Neuroscience} 31(13): 4811--4820.

\bibitem[{Brown et~al.(2014)Brown, Zeidman, Smittenaar, Adams, McNab, Rutledge,
  and Dolan}]{Brown2014}
Brown, H.~R.; Zeidman, P.; Smittenaar, P.; Adams, R.~A.; McNab, F.; Rutledge,
  R.~B.; and Dolan, R.~J. 2014.
\newblock {Crowdsourcing for cognitive science - The utility of smartphones}.
\newblock \emph{PLoS ONE} 9(7).

\bibitem[{Cranmer et~al.(2020)Cranmer, Sanchez~Gonzalez, Battaglia, Xu,
  Cranmer, Spergel, and Ho}]{symbolic20}
Cranmer, M.; Sanchez~Gonzalez, A.; Battaglia, P.; Xu, R.; Cranmer, K.; Spergel,
  D.; and Ho, S. 2020.
\newblock Discovering symbolic models from deep learning with inductive biases.
\newblock \emph{Advances in Neural Information Processing Systems} 33.

\bibitem[{Dezfouli et~al.(2019{\natexlab{a}})Dezfouli, Ashtiani, Ghattas, Nock,
  Dayan, and Ong}]{Dezfouli2019b}
Dezfouli, A.; Ashtiani, H.; Ghattas, O.; Nock, R.; Dayan, P.; and Ong, C.~S.
  2019{\natexlab{a}}.
\newblock {Disentangled behavioral representations}.
\newblock Technical report.

\bibitem[{Dezfouli et~al.(2019{\natexlab{b}})Dezfouli, Griffiths, Ramos, Dayan,
  and Balleine}]{Dezfouli2019a}
Dezfouli, A.; Griffiths, K.; Ramos, F.; Dayan, P.; and Balleine, B.~W.
  2019{\natexlab{b}}.
\newblock {Models that learn how humans learn: The case of decision-making and
  its disorders}.
\newblock \emph{PLoS Computational Biology} 15(6): e1006903.

\bibitem[{Ellsberg(1961)}]{Ellsberg1961}
Ellsberg, D. 1961.
\newblock {Risk, ambiguity, and the Savage axioms}.
\newblock \emph{Quarterly Journal of Economics} 75(4): 643--669.

\bibitem[{Finn, Abbeel, and Levine(2017)}]{Finn2017}
Finn, C.; Abbeel, P.; and Levine, S. 2017.
\newblock {Model-agnostic meta-learning for fast adaptation of deep networks}.
\newblock \emph{34th International Conference on Machine Learning, ICML 2017}
  3: 1856--1868.

\bibitem[{Garnelo et~al.(2018)Garnelo, Rosenbaum, Maddison, Ramalho, Saxton,
  Shanahan, Teh, Rezende, and Eslami}]{Garnelo2018}
Garnelo, M.; Rosenbaum, D.; Maddison, C.~J.; Ramalho, T.; Saxton, D.; Shanahan,
  M.; Teh, Y.~W.; Rezende, D.~J.; and Eslami, S.~M. 2018.
\newblock {Conditional neural processes}.
\newblock Technical report.

\bibitem[{Gesiarz, Cahill, and Sharot(2019)}]{Gesiarz2019}
Gesiarz, F.; Cahill, D.; and Sharot, T. 2019.
\newblock {Evidence accumulation is biased by motivation: A computational
  account}.
\newblock \emph{PLOS Computational Biology} 15(6): e1007089.

\bibitem[{Gilovich, Griffin, and Kahneman(2002)}]{Gilovich2004}
Gilovich, T.; Griffin, D.; and Kahneman, D. 2002.
\newblock \emph{{Heuristics and Biases: The Psychology of Intuitive Judgment}}.
\newblock Cambridge University Press.

\bibitem[{Hunt et~al.(2016)Hunt, Rutledge, Malalasekera, Kennerley, and
  Dolan}]{Hunt2016}
Hunt, L.~T.; Rutledge, R.~B.; Malalasekera, W. M.~N.; Kennerley, S.~W.; and
  Dolan, R.~J. 2016.
\newblock {Approach-Induced Biases in Human Information Sampling}.
\newblock \emph{PLOS Biology} 14(11): e2000638.
\newblock \doi{10.1371/journal.pbio.2000638}.

\bibitem[{Hunt et~al.(2017)Hunt, Rutledge, Malalasekera, Kennerley, and
  Dolan}]{Huntdata}
Hunt, L.~T.; Rutledge, R.~B.; Malalasekera, W. M.~N.; Kennerley, S.~W.; and
  Dolan, R.~J. 2017.
\newblock {Data from: Approach-induced biases in human information sampling}.
\newblock \emph{Dryad, Dataset}
  \urlprefix\url{https://datadryad.org/stash/dataset/doi:10.5061/dryad.nb41c}.

\bibitem[{Huys et~al.(2011)Huys, Cools, Friedel, Heinz, Dolan, and
  Dayan}]{Huys2011}
Huys, Q.; Cools, R.; Friedel, M.; Heinz, A.; Dolan, R.; and Dayan, P. 2011.
\newblock {Disentangling the roles of approach, activation and valence in
  instrumental and Pavlovian responding}.
\newblock \emph{PLoS Comp Biol In, Press} 1--30.

\bibitem[{Kahneman and Tversky(1979)}]{Kahneman1979a}
Kahneman, D.; and Tversky, A. 1979.
\newblock {Prospect theory: An analysis of decision under risk}.
\newblock \emph{Econometrica} .

\bibitem[{Ma et~al.(2006)Ma, Beck, Latham, and Pouget}]{Ma2006}
Ma, W.~J.; Beck, J.~M.; Latham, P.~E.; and Pouget, A. 2006.
\newblock {Bayesian inference with probabilistic population codes}.
\newblock \emph{Nature Neuroscience} 9(11): 1432--1438.

\bibitem[{Marr(1982)}]{Marr1982}
Marr, D. 1982.
\newblock \emph{{Vision: a computational investigation into the human
  representation and processing of visual information.}}
\newblock ISBN 0716712849.

\bibitem[{Mkrtchian et~al.(2017)Mkrtchian, Aylward, Dayan, Roiser, and
  Robinson}]{Mkrtchian2017}
Mkrtchian, A.; Aylward, J.; Dayan, P.; Roiser, J.~P.; and Robinson, O.~J. 2017.
\newblock {Modeling Avoidance in Mood and Anxiety Disorders Using Reinforcement
  Learning}.
\newblock \emph{Biological Psychiatry} 82(7): 532--539.

\bibitem[{Nickerson(1998)}]{Nickerson1998}
Nickerson, R. 1998.
\newblock {Confirmation bias: A ubiquitous phenomenon in many guises.}
\newblock \emph{Review of General Psychology} 2(2): 175--220.

\bibitem[{Pariser(2011)}]{Pariser2011}
Pariser, E. 2011.
\newblock \emph{{The filter bubble: how the new personalized Web is changing
  what we read and how we think}}.
\newblock Penguin.

\bibitem[{Ratcliff and McKoon(2008)}]{Ratcliff2008}
Ratcliff, R.; and McKoon, G. 2008.
\newblock {The diffusion decision model: Theory and data for two-choice
  decision tasks}.
\newblock \emph{Neural Computation} 20(4): 873--922.

\bibitem[{Rutledge et~al.(2015)Rutledge, Skandali, Dayan, and
  Dolan}]{rutledge2015dopaminergic}
Rutledge, R.~B.; Skandali, N.; Dayan, P.; and Dolan, R.~J. 2015.
\newblock Dopaminergic modulation of decision making and subjective well-being.
\newblock \emph{Journal of Neuroscience} 35(27): 9811--9822.

\bibitem[{Sharot(2011)}]{Sharot2011}
Sharot, T. 2011.
\newblock {The optimism bias}.

\bibitem[{Verma et~al.(2018)Verma, Murali, Singh, Kohli, and
  Chaudhuri}]{NDPS18}
Verma, A.; Murali, V.; Singh, R.; Kohli, P.; and Chaudhuri, S. 2018.
\newblock Programmatically interpretable reinforcement learning.
\newblock \emph{arXiv preprint arXiv:1804.02477} .

\bibitem[{Yu and Dayan(2005)}]{Yu2005}
Yu, A.~J.; and Dayan, P. 2005.
\newblock {Uncertainty, neuromodulation, and attention.}
\newblock \emph{Neuron} 46(4): 681--92.

\end{thebibliography}

\end{document}